\documentclass[sigconf]{aamas} 

\usepackage[procnumbered,linesnumbered,ruled,vlined]{algorithm2e}
\usepackage{amsmath}
\usepackage{caption}
\usepackage{subcaption}
\usepackage{lipsum}  
\usepackage{hyperref}
\usepackage{multirow}
\usepackage{multicol}

\newcommand*{\argmax}{\operatornamewithlimits{argmax}\limits}

\SetKw{Continue}{continue}
\newcommand{\ind}{\perp \!\!\! \perp}

\SetKw{KwGoTo}{go to}

\usepackage{etoolbox}

\makeatletter
\AtBeginEnvironment{procedure}{\let\c@algocf\c@procedure}
\makeatother

\newtheorem{theorem}{Theorem}[section]

\newtheorem{lemma}[theorem]{Lemma}

\theoremstyle{definition}
\newtheorem{definition}{Definition}[section]

\SetAlFnt{\small}
\SetAlCapFnt{\small}
\SetAlCapNameFnt{\small}
\usepackage{algorithmic}
\algsetup{linenosize=\tiny}

\usepackage{balance} 

\setcopyright{ifaamas}
\acmConference[AAMAS '21]{Proc.\@ of the 20th International Conference on Autonomous Agents and Multiagent Systems (AAMAS 2021)}{May 3--7, 2021}{Online}{U.~Endriss, A.~Now\'{e}, F.~Dignum, A.~Lomuscio (eds.)}
\copyrightyear{2021}
\acmYear{2021}
\acmDOI{}
\acmPrice{}
\acmISBN{}

\acmSubmissionID{349}

\title[AAMAS-2021 Formatting Instructions]
{Accelerating Recursive Partition-Based\\ Causal Structure Learning}
\author{Md Musfiqur Rahman\texorpdfstring{$^1,$}, Ayman Rasheed\texorpdfstring{$^1,$}, Md. Mosaddek Khan\texorpdfstring{$^1,$}, \and Mohammad Ali Javidian\texorpdfstring{$^2,$}, Pooyan Jamshidi\texorpdfstring{$^3,$}, Md. Mamun-Or-Rashid\texorpdfstring{$^1$}.}
\affiliation{
  \institution{\texorpdfstring{$^1$}, Department of Computer Science and Engineering, University of Dhaka\\
  \texorpdfstring{$^2$}, School of Electrical and Computer Engineering, Purdue University\\
  \texorpdfstring{$^3$}, Department of Computer Science and Engineering, University of South Carolina}
  }
\email{musfiq14shohan@gmail.com, aymanrasheed7@gmail.com, mosaddek@du.ac.bd,}
\email{mjavidia@purdue.edu, pjamshid@cse.sc.edu, mamun@cse.du.ac.bd}

\begin{abstract}
Causal structure discovery from observational data is fundamental to the causal understanding of autonomous systems such as medical decision support systems, advertising campaigns and self-driving cars. This is essential to solve well-known causal decision making and prediction problems associated with those real-world applications. Recently, recursive causal discovery algorithms have gained particular attention among the research community due to their ability to provide good results by using Conditional Independent (CI) tests in smaller sub-problems. However, each of such algorithms needs a refinement function to remove undesired causal relations of the discovered graphs. Notably, with the increase of the problem size, the computation cost (i.e., the number of CI-tests) of the refinement function makes an algorithm expensive to deploy in practice. This paper proposes a generic causal structure refinement strategy that can locate the undesired relations with a small number of CI-tests, thus speeding up the algorithm for large and complex problems. We theoretically prove the correctness of our algorithm. We then empirically evaluate its performance against the state-of-the-art algorithms in terms of solution quality and completion time in synthetic and real datasets.
\end{abstract}


\newcommand{\BibTeX}{\rm B\kern-.05em{\sc i\kern-.025em b}\kern-.08em\TeX}

\begin{document}
	%
	%
	\pagestyle{fancy}
	\fancyhead{}
	%
	%
	\maketitle 
	
	
	\section{Introduction}
	
	Causal structure discovery has emerged as a powerful computational method of identifying causal relationships from large quantities of data. Unlike the state-of-the-art statistical learning approaches, causal discovery examines the data generation procedure instead of inspecting the joint distribution of observed variables. Understanding and predicting causality in such a way 
	has received significant interest in a large number of real-life application such as gene regulatory network~\cite{yoo2002discovery, ellis2008learning}, advertising campaign~\cite{cai2013sada}, causal feature selection~\cite{aliferis2010local}, self-driving cars~\cite{kim2017interpretable}, medical decision support systems~\cite{constantinou2015causal} and many more besides~\cite{rashik2020speeding,Mahmud_2020}.
	
	\par Causal discovery is generally formulated as a probabilistic graphical model (i.e., causal graph) where each edge represents the causal relationship between variables~\cite{javidian2020learning,javidian2020hypergraph,javidian2020amp,javidian2019order}. If controlled experiments are not possible, inferring the causal relations becomes challenging. Constraint-based methods can identify these relations by exploiting conditional independence tests (CI-tests)~\cite{fukumizu2008kernel} when experiment samples are difficult to manipulate. 
	In a CI-test, when two variables are independent given a conditioning set (i.e., d-separated by the conditioning set, see definitions), then we can conclude that there is no direct connection (i.e. causal relation) between them. This helps to disconnect the variables during the construction of the causal graph. 
	Thus, following the faithfulness assumption (i.e., directed edges indicate causal relations)~\cite{koller2009probabilistic}, existing constraint-based methods such as IC-algorithm~\cite{radermacher1990probabilistic} and PC algorithm~\cite{spirtes2000causation} can discover partially directed acyclic graphs~\cite{pearl2009causality}. 

	\par Now, constraint-based methods find it very difficult to deal with d-separators in large problems. When the number of variables increases, the number of possible conditioning set grows exponentially, making the exploration of d-separators~\cite{pearl2009causality,cai2013causal} computationally expensive. Another major challenge is, as the size of the conditioning set grows (i.e., high-order) with the number of variables, CI-tests become unreliable and may experience Type II errors (i.e., false CI hypothesis is accepted)~\cite{Zhang2011,doran2014permutation,cai2013sada}. As a result, we observe (i) some edges go missing, although they should exist, (ii) some undesired false edges are introduced.

	To deal with these issues, researchers concentrated on recursive split-and-merge strategies~\cite{cai2013sada,geng2005decomposition, liu2017new, xie2008recursive, xie2006decomposition}. These methods divide the original variable set into multiple subsets such that each subset can be solved recursively as a sub-problem by using the existing causal discovery algorithms. The results of these sub-problems are later merged to recover the causal graph corresponding to the main variable set. To learn the causal graphs, in place of using constraint-based algorithms, we can employ these recursive approaches since they avoid  redundant CI-tests~\cite{zhang2019recursively}. In effect, they provide more accurate result in less amount of time.

	One notable algorithm named SADA is proposed in~\cite{cai2013sada}, which searches for causal cut over the variables in a sparse causal structure. Thus, it enables an efficient partitioning of the variables into subsets, and as such, produces a causal graph with a smaller number of samples. Nevertheless, SADA violates the d-separation (see Definition~\ref{def:d-separation}), and the causal cut searching process is costly since they are generated randomly and repeatedly for getting better decomposition. To address these issues, a new partitioning scheme, CAPA, decomposes the original variable set into three smaller subsets, and can operate with low-order CI-tests to detect more causal directions than its predecessors~\cite{zhang2019recursively}. However, the CAPA algorithm experiences severe execution time overhead, as well as a large number of CI-tests are required due to an additional partition (i.e., the third one). Recently, a split-and-merge strategy, named CP, has been developed that is reported to keep the run-time and the number of CI-tests at a lower margin. These are obtained by dividing the variable set into two partitions and executing a refinement process during the merge phase. This process is used to remove undesired false edges~\cite{yan2020effective}. It is worth noting that this process takes a major portion of the execution time and a large amount of CI-tests since CP cannot distinguish false edges separately and conducts CI-tests for all of them.

 	\par 	
	As far as we know, no previous research has investigated false edge detection due to the d-separation violation.
	This paper proposes a recursive  algorithm named \textbf{Dsep-CP} (d-separation preserving Causal Partition) to improve the scalability of causal discovery without sacrificing the solution quality.
	The first contribution of our paper includes the exploration of the conditions of d-separation violation  and the graphical structure analysis for detecting the false edges.
	{\sf Dsep-CP} employs this knowledge to refine the false edges of causal structure with a parsimonious number of CI-tests.
	This is the second key contribution of our work. 
	Our theoretical analysis proves that our algorithm returns the correct causal structure under reliable CI-tests. %
	Our empirical evaluation illustrates a substantial reduction in execution time up to 14\% compared to the state-of-the-art without sacrificing the solution quality. 
	In addition, during the refinement phase of our approach, it reduces the number of CI-tests up to 86\% in different experimental settings.

	
	\section{Background\:and\:Problem\:definition}
	
	This section formulates the causal discovery problem and the discuss the background necessary to understand our proposed method.

	\par A causal graph is a directed acyclic graph (DAG), $G=(V,E)$ where $V=\{v_1, v_2,\dots,v_n\}$ is a set of $n$ variables\footnote{Throughout this paper, we use the terms variables and nodes interchangeably.} and $E$ is a set of directed edges. Each edge $u \rightarrow v$ represents the cause and effect relations between variables $u$ and $v$ where $u$ is the parent (cause) and $v$ is the child (effect) in the DAG. 
	Let $D=\{x_1,x_2,\dots,x_m\}$ represents a data sample set where each sample is a vector, i.e., $x_i = \{x_{i1}, x_{i2},\dots,x_{in}\}$. Here $x_{ij}$ defines the value of variable $v_j$ in sample $x_i$. 
	The sample data $D$ is generated from a causal graph $T$ with $n$ nodes (we define it as a true graph). We consider that during the generation of data samples, the Causal Sufficiency~\cite{cai2013sada} (i.e., the non-existence of latent confounders of any two observed variables) and the Faithfulness condition ~\cite{koller2009probabilistic} (i.e., directed edges indicating causal relations) assumptions are followed.
	However, we aim to recover the causal graph $G^*$ from $D$ (Equation~\ref{eqn:causal_graph}). Here $score(G,T)$ indicates how close $G$ is compared to $T$ with respect to the causal relations between variables. Ideally, $G^*$ is identical to $T$, if not, as close as possible. However, finding exact $T$ is computationally infeasible due to Markov equivalence classes~\cite{chickering2002learning} (i.e., different graphs with the same conditional independence relations among the variables). 
	
				\begin{equation}
	\label{eqn:causal_graph}
	G^* = \argmax_{G} Score(G,T)
	\end{equation}

	\theoremstyle{definition}
	\begin{definition}[D-separation]
		\label{def:d-separation}
		Two variables $u$ and $v$ are called d-separated with respect to a conditioning set $Z$ if at least one of the following two conditions hold: $(i)$ the path (i.e., a consecutive sequence of edges) between $u$ and $v$ contains a mediator $(u \rightarrow \textbf{w} \rightarrow v)$ or a confounder $(u \leftarrow \textbf{w} \rightarrow v)$ where $w\in Z$, $(ii)$ the path between $u$ and $v$ contains a collider $(u \rightarrow \textbf{w} \leftarrow v)$ where $w$ and its descendants are not in $Z$ ~\cite{pearl2000models}. 
		We employ CI-tests to determine d-separation\footnote{We use the terms conditional independence and d-separation interchangeably.} in DAGs utilizing the sample set $D$.
		During causal partitioning, we denote that d-separation is preserved if (i) adjacent variable pairs cannot be d-separated, (ii) non-adjacent variable pairs are either already divided into different subsets or d-separable in at least one of the subsets~\cite{zhang2019recursively}.
		Otherwise, d-separation is violated during that partitioning process. Notably, if two variables $u$ an $v$ are conditionally independent given a variable set $Z$, we represent it with $u\ind v|Z$. 
	\end{definition}
	
\begin{figure}[t]
		\centering
			\vspace{-4mm}
		\includegraphics[scale=0.5]{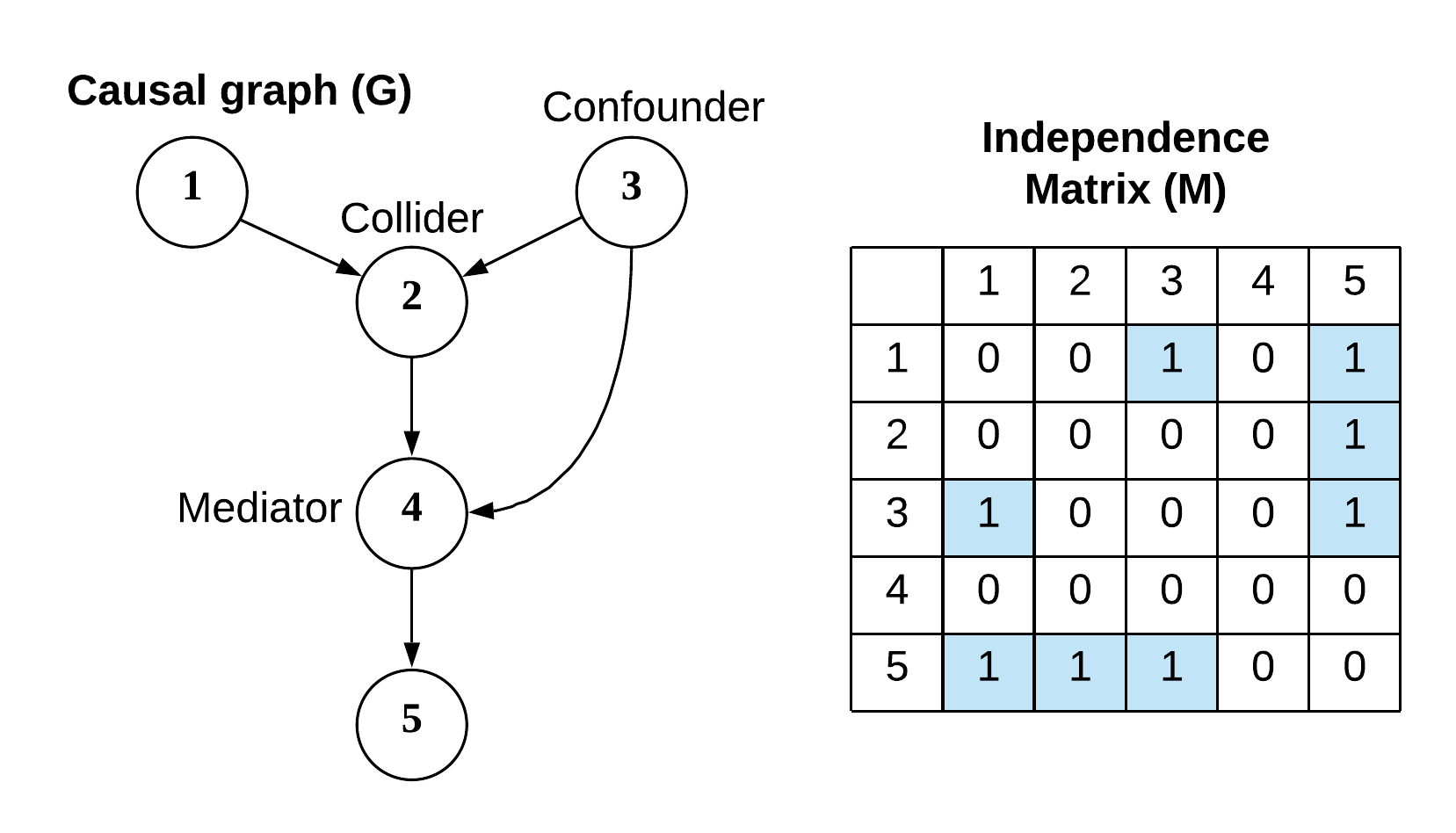}
		\vspace{-10mm}
		\caption{Y-structure causal graph $G$, matrix $M$ of $k\_order = 1$.}
		\label{Fig:example_graph}
	\end{figure}

	\theoremstyle{definition}
	\begin{definition}[Y-structure and Independence matrix]
		We define the formation of nodes as \textbf{Y-structure} where a collider exists with its parents and descendants.
		Whereas, $M$ denotes an independence matrix. Here, $M_{ij}=1$ indicates that $v_i \ind v_j |Z$ for some
		$Z \subseteq V\setminus \{v_i,v_j\}$ and $|Z| \leq k\_order$. Otherwise, they are not yet d-separated for $k\_order$ conditioning set. Here, $k\_order$ indicates the size of the conditioning set, and it increases from 0 to $k\_thresh$ which is a pre-specified maximum limit. Figure ~\ref{Fig:example_graph} shows an example of Y-structure causal graph $G$ and independence matrix $M$ of $k\_order = 1$. It can be seen from this figure that in $G$, variable $2$ is a collider for variables $1$ and $3$, variable $3$ a confounder for variables $2$ and $4$, variable $4$ a mediator for variables $2$ and $5$ according to Definition~\ref{def:d-separation}.
		Here, $1 \ind 3 | \{\}$, $1 \ind 5 | \{4\}$, $2 \ind 5 | \{4\}$ and $3 \ind 5 | \{4\}$ are represented\:in\:$M$.
		\label{def:y-structure}
	\end{definition}
	\setlength{\textfloatsep}{3mm}
	\setlength{\floatsep}{3mm}
		\vspace{-6mm}
	\subsection{Causal Refinement}

	During the merging phase of SADA, in order to hold the acyclicity assumption (i.e., the causal graph is a DAG), it removes the less significant edge(s) (according to CI-test) when a cycle is found.
    Whereas, if two paths are found between a pair of variables, it eliminates specific edges with CI-tests. However, SADA fails to preserve d-separation during the partitioning phase. As a result, a number of redundant edges is created. To resolve this issue, CAPA splits the problem into three smaller sub-problems. The main purpose of the third partition is to ensure the preservation of d-separation. Due to this third sub-problem, CAPA does not have to perform any refinement during the merging phase. However, this third partition created during each recursive decomposition notably degrades CAPA's runtime performance.

	The CP algorithm first arranges the variables 
	in descending order by the number of non-adjacent variables according to the independence matrix $M$. This strategy makes the partitioning process more efficient and extremely fast~\cite{yan2020effective}. 
	After that, CP divides the variable set $V$ into three non-overlapping sets $A, B$ and $C = V\setminus(A \cup B)$ such that $\forall _{i\in A}, \forall _{j\in B}, M_{ij}=1$. 
	Based on this decomposition, this algorithm creates two partitions $V_1= A \cup C , V_2=B\cup C$. However, in case of unsuccessful decomposition, CP increases the $k\_order$ by 1 and restarts the partitioning method.
	In the merge phase of CP, it combines the resulting graphs from its two sub-problems and executes a refinement procedure.
	During the causal partitioning, the d-separation can be violated in some cases, which may produce false edges. The aim of the refinement step is to remove those edges from the causal graph. Therefore, CP performs CI-tests for every pair of variables conditioning on their parents.
	
	In light of the above discussion, to develop an efficient causal discovery algorithm, we have to address two challenges: (i) detect the false edges due to the d-separation violation, (ii) provide a strategy that removes those false edges efficiently without losing the solution quality. In the following section, we describe our method that deals with the above challenges.

	\section{The Dsep-CP Algorithm}
	{\sf Dsep-CP} is a recursive method that effectively performs scalable causal discovery. The algorithm initiates with an original variable set, and at each level of recursive decomposition, the algorithm performs three principal operations. Initially, the \textbf{Find Causal Partitions} step takes place to find two causal partitions of the variable set, being careful about the d-separation violation. Secondly, the \textbf{Recursive Dsep-CP} calling step proceeds to solve the two sub-problems corresponding to the two partitions. Finally, the \textbf{Dsep-CP Refinement} step refines the causal graph that is discovered and merged from the two partitions. 
	
	\par Algorithm~\ref{algo1} shows the pseudo-code for {\sf Dsep-CP}. The input of {\sf Dsep-CP} is the variable set $V$ for which we need to construct the causal graph. The algorithm starts by checking the size of the variable set. If the size of the variable set is smaller than (or equal to) a suitable threshold, $graph\_thresh\_size$, the PC algorithm
	\footnote{We have followed CP~\cite{yan2020effective}. Other constraint-based  algorithms can also be used.} 
	is run on this variable set for constructing causal relations (Algorithm 1: lines 1-3). Next, at line 4, the Find Causal Partitions function is called for finding suitable partitions of the main variable set.
	%
	%
	\begin{algorithm}
		\DontPrintSemicolon

		\SetKwInOut{Input}{Input}  
		\Input{A finite set of variables $V=\{v_1, v_2, \ldots, v_n\}$\\
			
		}

		\KwOut{Discovered causal structure, G}
		
		\If{$len(V) \leq graph\_thresh\_size$} {
			$G \gets  $ PC\_Algorithm(V) \hspace{10mm}// Discover causal relation  \;
			
			\Return{ $G$}\;
		}

		$[V_1, V_2]  \gets  $   Find-Causal-Partitions$(V)$
		
		\If{$\max(len(V_1), len(V_2)) = len(V)$}{
			$G \gets  $ PC\_Algorithm(V) \hspace{10mm}// Discover causal relation  \;
			\Return{ $G$}\;
		}
		
		\Else{
			$ graph_1 \gets $Dsep-CP($V_1$) \;
			$ graph_2 \gets $Dsep-CP($V_2$) \;
		}
		
		$G^{\prime} \gets$ Merge$(graph_1, graph_2)$ 
		\hspace{5mm}// Resolving conflicts\;
		$G \gets$  Dsep-CP-Refinement$(G^{\prime}, graph_1, graph_2)$
		
		\Return{ $G$}\;
		\caption{ {\sf Dsep-CP}(V) }
		\label{algo1}
	\end{algorithm}
	\par 
	The \textbf{Find Causal Partitions }step (Procedure~\ref{procedure1}) starts with initializing partitioning sets $A,B,C$ as empty sets, independence matrix $M $ (see Definition~\ref{def:y-structure}) as a $|V| \times |V|$ size zero matrix 
	and $k\_order$ as 0 (line 1).
	Next, the matrix $M$ is calculated at lines 2-3 and $M$ is copied to $M^{\prime}$ at line 4.
	In the next step, we partition $V$ into three non-overlapping variable sets $A,B$, and $C$ according to the optimization process used in CP~\cite{yan2020effective}. In that process (lines 5-13), we execute $|V|$ iterations, and in each iteration, we choose variable $w$ with the highest priority and assign it in either $A, B$, or $C$ .
	At line 6, we select the variable $w$ which is d-separated from the most number of variables according to $M^{\prime}$ (a copy of $M$), i.e., $w=\argmax_{r\in\{1,\dots,|V|\}} \sum_{c=1}^{|V|}
	M^{\prime}_{c,r} $.
	 The variable $w$ is either appended to the set $B$ (or $A$) if it is independent of all the variables existing in 
	$A$ (or $B$) (lines 7-10).
	 Otherwise, it is added into C (line 12). At last, the $M^{\prime}$ matrix is updated for $w$ (line 13) by setting the column values of $w$ to zero so that other variables with lower priority can be chosen in the next iteration at line 6.
	 \par
\begin{figure}[t]
		\centering
			\includegraphics[scale=0.4]{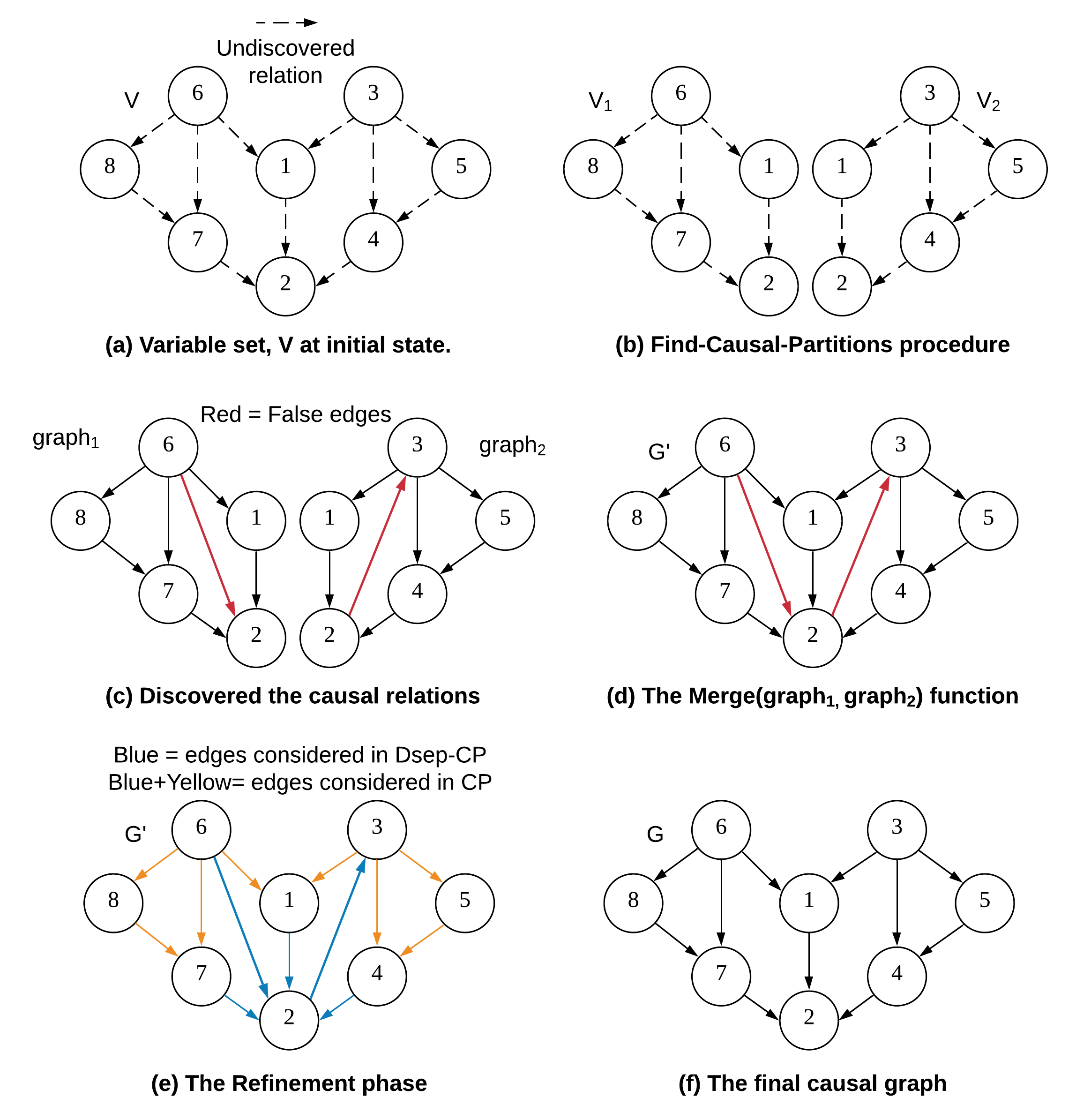}
			\vspace{-4mm}
		\caption{
			Illustration of Dsep-CP algorithm. 
			Find-Causal-Partitions procedure divides the variable set $V$ of Figure~\ref{algo-simulation}a into $V_1$=  \{1, 2, 6, 7, 8\} and $V_2$ = \{1, 2, 3, 4, 5\} in Figure~\ref{algo-simulation}b. Since, partition size has reached the size threshold (=5), we can execute the causal discovery PC algorithm. It discovers $graph_1$ and $graph_2$ in Figure~\ref{algo-simulation}c.
			In Figure~\ref{algo-simulation}d, the merge function combines the discovered graphs to form $G$. We observe that, $(6\rightarrow2)$ and $(2\rightarrow3)$ are false edges.
			The	refinement phase performs CI-test for 5 (blue) edges=\{(1,2), \textcolor{red}{(2,3)}, (4,2), \textcolor{red}{(6,2)}, (7,2)\} and refines 2 edges in Figure~\ref{algo-simulation}e.
			Figure~\ref{algo-simulation}f is the final graph.
		}			
		\label{algo-simulation}
    	\end{figure}

	After finishing all iterations, if
	the partition $C$ is not smaller than the other two partitions $A$ and $B$, we do not consider these as efficient partitions and try again with higher $k\_order$ if it is less than the allowed max value, $k\_thresh$.
	In this case, the procedure increases $k\_order$ by one and goes to line~\ref{outer_loop} to start the partitioning procedure again (Procedure~\ref{procedure1}: lines 14-16). 
	The procedure now updates $M$ for many variable pairs that could not be d-separated earlier but becomes independent now conditioning on a higher-order variable set.
	 On the contrary, in case of efficient partitioning, lines 17-19 merge $A$ and $C$ to form $V_1$ and also merge $B$ and $C$ to form $V_2$. Finally, these two partitions are returned from Procedure~\ref{procedure1}, and we use them in the next steps of Algorithm~\ref{algo1}.
	For example, we partition the variables of Figure~\ref{algo-simulation}a into three sets $A=\{6, 7, 8\}, B=\{3, 4, 5\}$ and $C=\{1, 2\}$ for $k\_order=0$. Then we take the union of $A$ and $C$ to get $V_1=\{1, 2, 6, 7, 8\}$ and the union of $B$ and $C$ to get $V_2=\{1, 2, 3, 4, 5\}$ (Figure~\ref{algo-simulation}b).
	Here, the dashed directed edges indicate the causal relations between variables that are not yet discovered but exist in the true graph (details in Section 2).
		\begin{procedure}[t]
		\DontPrintSemicolon 
		\SetKwInOut{Input}{Input}  
		\Input{A finite set of variables $V=\{v_1, v_2, \ldots, v_n\}$
		}
		\KwOut{The causal partitions $V_1, V_2$}
		
		Initialize $A,B,C$ as empty set, M as $|V|\times|V|$ zero matrix and $k\_order \gets 0$ \\
		
		\ForEach{$v_i,v_j  \in V$}{ \label{outer_loop}
			$M_{i,j} \gets 1$ such that  $v_i \ind v_j|Z$ where
			$\exists Z \subseteq V\setminus\{v_i,v_j\}$ and $|Z|\leq k\_order $
		}
		
		$M^{\prime}= Duplicate(M)$  \\
		\For{$iteration \gets 1$ \textbf{to} $|V|$}{
			
			$w \gets  \argmax_{r\in\{1,\dots,|V|\}} \sum_{c=1}^{|V|}
			M^{\prime}_{c,r} $

			\uIf{$w$ independent of $\forall v_i \in A$ according to $M$}{
				$B \gets Append(B,w)$}
			\uElseIf{$w$ independent of $\forall v_i \in B$ according to $M$}{
				$A \gets Append(A,w)$}
			\Else
			{$C \gets Append(C,w)$}
			Update $M^{\prime}$ for $w$ \hspace{3mm}// $M^{\prime}$ gets updated, M remains fixed \;
			
		}

		\If{ $(|C| \geq |A| +|B|) \land  (k\_order+1 < k\_thresh)  $}
		{
			$k\_order \gets	 k\_order + 1$ \\
			\textbf{Goto} line \ref{outer_loop}
			\hspace{10mm}// Try again for efficient decomposition \;
		}
		\Else{
			$V1 \gets A\cup C$ 	\hspace{10mm}// Two new partitions\;
			$V2 \gets B\cup C$
		}
		
		\Return{$[V_1 , V_2]$}\;
		\caption{Find-Causal-Partitions($V$)}
		\label{procedure1}
	\end{procedure}
	\par In Algorithm~\ref{algo1} (lines 5-7), if the largest of the resulting two partitions $V_1$ and $V_2$ is the same as the main variable set $V$, then 
	it indicates that the partitions are not efficient even after the partitioning procedure. This situation occurs when the variables are more connected to each other and may form a dense sub-graph.
	In that case, we execute the \textbf{PC\_Algorithm(V)} for constructing the causal structure G, similar to Algorithm~\ref{algo1}: line 2.
	The PC algorithm discovers and stores the causal relations in an adjacency matrix $G.dir$. Here $G.dir_{i,j} =1 $ indicates a directed edge, $v_i\rightarrow v_j$ and $0$ means no edge exists. 
	 We use dot operation to express specific attributes of a graph, in this case, the direction matrix of graph $G$.
	It is worth noting that our algorithm is compatible to deal with both directed and undirected causal relations. 
	So, if the PC algorithm cannot detect the directions properly, then our algorithm considers the causal structure as an undirected causal skeleton. Under this circumstance, {\sf Dsep-CP} is still able to perform  its improved refinement procedure and provide significant result.
	Therefore, if the causal skeleton is undirected then $G^{\prime}.dir_{i,j} =1 $ represents only the existence of an edge.
\par
	In Figure~\ref{algo-simulation}b, we consider $graph\_thresh\_size = 5$ and thus, we execute the PC algorithm on each partition for discovering the causal structures.
	The resulting graphs are shown in Figure~\ref{algo-simulation}c.
	Here, black edges indicate the relations that are correctly discovered, and red edges refers to false edges that are created due to d-separation violation (see Definition~\ref{def:d-separation}). 
	The discovered causal relations
	$\{(1\rightarrow2), (6\rightarrow1), (6\rightarrow2), (6\rightarrow7), (6\rightarrow8), (7\rightarrow2), (7\rightarrow8)\}$ form $graph_1$ from the left partition and $graph_2$ consists of $\{(1\rightarrow2), (2\rightarrow3), (3\rightarrow1), (3\rightarrow4), (3\rightarrow5), (4\rightarrow2), (5\rightarrow4)\}$ produced from the right partition. %

%
	\par
	Notably, the d-separation may get violated in some specific situations during the partitioning process.
	For example, in the left partition $V_1$ of Figure~\ref{algo-simulation}b, to detect the causal relation between 2 and 6, we perform CI-tests conditioning on Z=\{1,7,8\}. The result of the CI-tests indicates that Z cannot d-separate variables 2, 6 and creates the edge $6\rightarrow2$ in Figure~\ref{algo-simulation}c. However, this is a false edge since it does not exist in the true graph of $V_1$ in Figure~\ref{algo-simulation}b.
	In the whole true graph of Figure~\ref{algo-simulation}a, the paths $6\rightarrow7\rightarrow2$ and $6\rightarrow1\rightarrow2$ can be blocked by conditioning on 7 and 1, respectively.
	However, conditioning on collider node 1 of the figure also opens the paths $6\rightarrow1\rightarrow3\rightarrow4\rightarrow2$  and $6\rightarrow1\rightarrow3\rightarrow5\rightarrow4\rightarrow2$. Hence, to make 6 and 2 conditionally independent, we have to condition on \{1,3,7\} or \{1,4,7\}.
	But we do not have access to all of these d-separators from the left partition, $V_1$ in Figure~\ref{algo-simulation}b. Hence, it can be inferred that d-separation is violated, and a false edge $6\rightarrow2$ is created. Similarly, another false edge $2\rightarrow3$ is constructed. 
	

	\begin{procedure}[t]
		\DontPrintSemicolon 
		\KwIn{Merged graph $G$, sub-problem graphs $graph_1,graph_2$}
		\KwOut{Refined graph $G$}
		
		\If{G structure is directed}{  					 \label{proc21}
			$collider\_set \gets get\_colliders(G)$		 \label{proc22}
		}
		\Else{											\label{proc23}
			\ForEach{$v_k \in $ variables in G}{		\label{proc24}
			
				$collider\_set \gets Append(collider\_set,v_k)$ in case $\forall v_i, \forall v_j \in neighbors(v_k) $ : $ \exists Z, v_i \ind v_j| Z$ and $v_k \notin Z $							\label{proc25}
			}
			
		}
		
		\ForEach{ $collider \in collider\_set$}{  			\label{proc26}
			
			\If{$ collider \notin graph_1$ or $collider \notin graph_2$}{
				\label{proc27}
				\textbf{Goto} line~\ref{proc26} 	\label{proc28}
				\hspace{10mm}// Skip this collider\;
			}

			$neighbor\_set \gets get\_neighbors(collider)$  \label{proc29} \\
			
			\If{G structure is directed}{		\label{proc211}
				$parent\_set \gets get\_parents(collider )$  \label{proc212}
			}
			\Else{		 \label{proc213}
				$parent\_set \gets neighbor\_set $  \label{proc214}
			}

			\If{$collider$ not in  $Y-structure$ }
			{ \label{proc216}
			\textbf{Goto} line~\ref{proc26}   \label{proc217}
			}

			\ForEach{$cur\_par \in neighbor\_set$}{	\label{proc210}

				Remove the edge between $collider$ and $cur\_parent$ of $G$
				if $\exists Z \subseteq parent\_set ( 0<|Z|< k\_order\_thresh ) $ such that $collider \ind cur\_parent| Z$ 
				 \label{proc218}
			}
		}
		\Return{$G$}\;		\label{proc219}
		\caption{Dsep-CP-Refinement($G, graph_1, graph_2$)}
		\label{procedure2}
	\end{procedure}
	%
	%
	%
	%
    	
	\par In case of efficient partitions at Algorithm~\ref{algo1}: line 4, we further proceed and recursively run the {\sf Dsep-CP} algorithm for each individual partitions $V_1,V_2$ (lines 9-10).
	The causal graphs constructed at lines 2 and 6 are returned as the return value of the sub-procedures called at line 9 and line 10.
	The resulting causal graphs, $graph_1$ and $graph_2$,  are combined with a $Merge(graph_1,graph_2)$ function to form $G^{\prime}$ at Algorithm 1: line 11.
	In this function, a new causal structure $G^{\prime}$ is created appending the variables of both graphs i.e., $G^{\prime}.V = graph1.V + graph2.V $. 
	We merge another attribute, direction matrix, $dir$ such that $G^{\prime}.dir_{i,j} = 0$ when any of the graphs has $dir_{i,j}=0$. 
	Otherwise, $G^{\prime}.dir$ takes the same values from both graphs. Precisely, these conditions can be expressed with an AND operation in Equation~\ref{eqn:merge_dir}.
	\begin{eqnarray}
	\label{eqn:merge_dir}
	G.dir_{i,j} = graph_1.dir_{i,j} \land graph_2.dir_{i,j}
	\end{eqnarray}
	The reason is that there can be some situations when we observe conflicting relations between variables in the graphs produced from two partitions. In one partition, two variables may be causally connected, and in another partition, they may be disconnected. In that case, we take the independent relation, $v_i \ind v_j$ (i.e., disconnected) as the correct one. Because, if two variables $v_i$ and $v_j$ are found independent in one partition, they are still independent after merging the sub-problems since the conditioning set also exists in the merged graph.
	The causal graph of Figure~\ref{algo-simulation}d is the result of merging two graphs produced in Figure~\ref{algo-simulation}c. This resulting graph $G$ has 13 edges in total: $\{(1\rightarrow2), (2\rightarrow3), (3\rightarrow1), (3\rightarrow4), (3\rightarrow5), (4\rightarrow2), (5\rightarrow4) (6\rightarrow1), (6\rightarrow2), (6\rightarrow7), (6\rightarrow8), (7\rightarrow2), (7\rightarrow8)\}$. In some cases, there exist some undesired false edges resulting from inefficient decomposition due to d-separation violations.
	Here, $(2\rightarrow3)$ and $(6\rightarrow2)$ are false edges according to the true graph of Figure~\ref{algo-simulation}a (see Section 2 for more details).
	\par The \textbf{ Dsep-CP\_Refinement} (Procedure~\ref{procedure2}) is called as a  sub-procedure of Algorithm~\ref{algo1} at line 12.
	Similar to the other algorithms described in Section 2, the purpose of the refinement procedure in {\sf Dsep-CP} is to refine the resulting causal graph constructed from the sub-problems. 
	This procedure finds a set of selected edges, $E$ and eliminates the false ones by executing CI-tests on each of them.
	To measure the performance of the refinement procedure, we define a function $Hit\_rate(E)$ in Equation~\ref{eqn:hit_rate} that indicates the proportion of the number of CI-tests required for refining the false edges with respect to the total number CI-tests performed on all edges in $E$.
	\begin{equation}
	Hit\_rate(E) = \frac{ F(E)}
	{F(E) +  T(E) }
	\label{eqn:hit_rate}
	\end{equation}
	\begin{equation}
	E^{*} = \argmax_{E \subseteq V\times V} \quad Hit\_rate(E)
	\label{eqn:maximize_hit}
	\end{equation}
	Here, $F(E)$ is the number of CI-tests requited to find the false edges in $E$. This is fixed for a specific causal structure. $T(E)$ is the number of CI-tests that are required to detect which edges are not false. Note that, these CI-tests do not help us to remove undesired edges. So, they are redundantly performed in a refinement procedure.
	The objective is to choose an optimal edge set $E^{*}$ which includes all the false edges and least number of edges that are not false such that the function $Hit\_rate$ is maximized in Equation~\ref{eqn:maximize_hit}.
	Notably, in the CP algorithm refinement, CI-tests needs to be performed on the whole edge set of the causal structure.
	 However, in {\sf Dsep-CP}, we reduce the size of $E$ and therefore, reduce the number of redundant CI-tests. This reduction is performed by searching for \textbf{Y-structure} (see Definition~\ref{def:y-structure}) and utilizing the colliders in it (correctness proved in Theorem~\ref{thm:collider}).
	 The colliders in the Y-structure must belong to both sub-graphs (i.e., $graph_1, graph_2$). It indicates the presence of the colliders in the variable set $C$ that was formed during the execution of the Find-Causal-Partitions procedure (Procedure~\ref{procedure1}: line 12).
	\par 
	To remove the false edges, the {\sf Dsep-CP}\_Refinement takes merged graph $G$ and the resulting graphs ($graph_1, graph_2$) from the sub-problems as input.
	The first step of this procedure is to create a $collider\_set$ that consists of colliders of the graph.
	If the resulting causal graph is directed, we find colliders by choosing such nodes that have at least two incoming edges (Procedure 2: lines~\ref{proc21}-\ref{proc22}). In Figure~\ref{algo-simulation}e, variables 1, 2, 4 and 7 are colliders, i.e., $collider\_set =\{1, 2, 4, 7\}$.
	On the contrary, if the graph is undirected, we first select a node $v_k$ and a list of its neighbors. Then for each pair of its neighbors $v_i,v_j$, we check if $v_i\ind v_j$ for any conditioning set $Z$. Now, if they are independent but the considered node $v_k$ does not belong to the set $Z$, we can admit it as a collider (Procedure 2: lines~\ref{proc23}-\ref{proc25})~\cite{spirtes2000causation}.
    For example, in Figure~\ref{algo-simulation}e, 
	we can consider about variable 2 and its neighbors = $\{1,3,4,6,7\}$. Now, $1\ind7|\{6\}$ but $2\notin \{6\}$. So, 2 is a collider in this graph.
	\par
	Next, we iterate through the colliders listed in $collider\_set$ and remove the suitable false edges (lines~\ref{proc26}-\ref{proc218}).
	 If a collider does not exist in both graphs, $graph_1$ and $graph_2$, we skip it (lines~\ref{proc27}-\ref{proc28}). 
	At line~\ref{proc29}, we save all the neighbors connected to the collider in $neighbor\_set$.
	After that, if the causal graph is directed, we acquire the list of collider's parents in the $parent\_set$ with the help of directed adjacency matrix $G.dir$.
	If not directed, then we use collider's $neighbor\_set$ as the alternative of  its $parent\_set$.
	(Procedure~\ref{procedure2}: lines~\ref{proc211}-\ref{proc214}).
	Next, lines~\ref{proc216}-\ref{proc217} checks whether the collider is a descendant of another collider that (i.e., the ancestor) has at least two of its parents in different sub-graphs ensuring the \textbf{Y-structure}. For example, in Figure~\ref{algo-simulation}e, we continue with variable 2 since its ancestor 1 (a collider) has two parents $6\in graph_1$ and $3\in graph_2$.
	\par 
	After that, we iterate through each $cur\_par$ in the $neighbor\_set$ of the collider (lines~\ref{proc210}-\ref{proc218}).
	We perform a conditional independence test between the collider and the $cur\_par$ where any subset (size less than pre-specified $k\_thresh$) of the $parent\_set$ can be used as conditioning set. 
	 If they are independent, we remove the edge between the collider and the $cur\_par$. 
	 Finally, after removing suitable edges following the conditions, the refined graph $G$ is returned from the {\sf Dsep-CP}-Refinement procedure.
	  The final causal graph is returned from Algorithm~\ref{algo1} at line 13. 
	We can visualize the illustration of these steps with Figure~\ref{algo-simulation}e. Our current $collider\_set =\{1, 2, 4, 7\}$. Now, we exclude those variables that do not belong to both graphs and are not descendant of any nodes in Y-structures. So, the updated $collider\_set$ is $\{2\}$. 
	For collider 2, we execute CI-tests on $\{(1,2), (2,3), (4,2), (6,2), (7,2)\}$. Altogether, we examine in total 5 distinct edges and among them two of them are removed i.e., $6\ind 2 | \{1,3,7\}$ and $2 \ind 3| \{1,4,6\}$. So, the hit rate = 2/5. On the other hand, a major drawback in CP algorithm is that, it tests every edge in the graph for refinement. 
	So, for this graph, CP performs CI-tests for all 13 edges and refines only 2 of them. This results in a hit rate of 2/13. 
	After the refinement step depicted in Figure~\ref{algo-simulation}e, we can see the final causal graph in Figure~\ref{algo-simulation}f.
\section{theoretical analysis}

	In this section, we prove the validity of {\sf Dsep-CP} in detecting the false edges created due to d-separation violation (see Definition~\ref{def:d-separation}). The violation occurs during the partitioning process when we construct partitions $A,B,C$ from $V$ and create two sub-problems $V_1 = A\cup C,V_2= B\cup C$. Notably, the following Lemma~\ref{lem:var_pos} builds the premise for proving the correctness of our refinement approach in Theorem~\ref{thm:loc_false}. Lemma~\ref{lem:par_cond} verifies the optimization technique used in this approach.  We also analyze the time complexity of our algorithm in this section.
	\begin{lemma}
		\label{lem:var_pos}
		If a false edge $x\rightarrow y$ created in sub-problem $V_1$ due to d-separation violation, can be removed  after merging in $V$ with independence test $x\ind y|\{w_{A\cup C},w_B\}$ then we can prove that there exist $x\in A, y\in C, w_{A\cup C}\in C\;and\; w_B\in B$.
		\begin{proof}
			We know that any pair of variables $x,y$ in $A$ are d-separable in $V_1=A\cup C$ if they are non-adjacent~\cite{yan2020effective}. So, the only d-separation violation can occur when $x\in A$ and $y\in C$. In the sub-problem $V_1$, $x \not\!\perp\!\!\!\perp y$ but $x\ind y|Z$ in the merged graph, therefore, it is obvious that few variables $w_B \subset Z$ must be from the other partition $B$. So, the rest, $w_{A\cup C}\subset Z$ is from set $A$ or $C$.
		\end{proof}
	\end{lemma}
\vspace{-2mm}
	\begin{theorem}
		\label{thm:loc_false}
		Existence of false edge $x\rightarrow y$ can be detected through Y-structures.
		\begin{proof}
			From Lemma~\ref{lem:var_pos}, we can infer that
			the variable $w_B$ blocks the causal effect from variable $x$ to $y$ that flows through partition B. However, we know that $x\ind w_B|c \subset C$ since $x\in A, w_B\in B$ are divided during the partitioning process. That means, the path between $x$ and $w_B$ gets open when we condition
			on a variable $w^{\prime}\in w_{A\cup C}$.
			This situation occurs only when we condition on variable $w^{\prime}$ to block the chain 
			$x\rightarrow w^{\prime}\rightarrow\dots\rightarrow y$ (path 1) and $w^{\prime}$ also acts as a collider in the path 
			$x\rightarrow w^{\prime}\leftarrow w_B\rightarrow\dots\rightarrow y$ (path 2). 
			Since, $w^{\prime}$ is a collider and a false edge $x\rightarrow y$  exists, path 1 must be a chain. 
			We cannot replace path 1  with another path $x\rightarrow w^{\prime}\leftarrow\dots\rightarrow y$, because in that case $x,y$ would be independent and no false edge would exist in the sub-problem.
			However, $y$ is a descendant of collider $w^{\prime}\in C$ and $w_B\in B$. So, the arrangement of the variables forms a $Y$ structure. That concludes that we can locate the possible false edges by searching for $Y$ structure without testing all the edges of the merged graph. As a result, if the PC algorithm and the CI-tests are both reliable, {\sf Dsep-CP} returns the actual causal graph by removing the false edges.
		\end{proof} 
		\label{thm:collider}
	\end{theorem}
	\vspace{-2mm}
	\begin{lemma}
		To check the independence between $x$ and $y$ in Y-structure, CI-tests with only conditioning on the parents of the collider $y$ is sufficient.
		\begin{proof}
			From Theorem~\ref{thm:loc_false}, we can locate the position of the variable $y$ that may have false edges but not sure about which of its connected edges is false. So, we have to perform CI-test for each of its edges. However, in Theorem~\ref{thm:loc_false}, we prove that the direction of the causal effect is from $x$ to $y$. Now, a path $y\rightarrow\dots\rightarrow x$ is not possible because it creates a cycle which violates the acyclicity assumption.
			It indicates that the causal effect of $x$ passes through the parents of $y$. Therefore, we can perform CI-test conditioning on only the parent variables of $y$ instead of conditioning on parents of both $x$ and $y$.
		\end{proof} 
		\label{lem:par_cond}
	\end{lemma}
	We now consider time complexity of our algorithm. During the partitioning procedure,
	 we use a conditioning set of size maximum $k\_order\_thresh  (=\sigma)$. So, in the worst case, we have to calculate the independence matrix $M$ (see Definition~\ref{def:y-structure}) for $\sigma\;order$. For variable set $V=\{v_1, v_2,\dots,v_n\}$ with conditioning set of size $k\_order=\{0,\dots,\sigma\}$, we have to perform $n^2*(\binom{n}{0}+\binom{n}{1}+\dots+\binom{n}{\sigma}) \approx n^2* n*n^\sigma = n^{\sigma+3}$ number of CI-tests to construct independence matrix $M$.
	During the refinement step, {\sf Dsep-CP} performs CI-tests only for those edges that are connected to any collider of the Y-structure and remain in the middle partition, $C$ (see Theorem~\ref{thm:collider} for the proof). If the size of the variable set is $|V|=n$, the size of final selected node set is, $n^{\prime} = n*d*p$ where, $d$ is the ratio of the variables in $C$ to $V$ and $p$ is the ratio of the Y-structure colliders in $C$ to all variables in $C$. Both of these ratios help lowering $n$.
	Therefore, if the average degree is $e$, the required number of CI-tests is 
	$(n^{\prime}*e) *(\binom{m}{0}+\binom{m}{1}+\dots+\binom{m}{\sigma}) \approx n* m^{\sigma+1}_{max}$ (Here, $m_{max}$ denotes the max number of neighbors). Finally, the complexity of the PC algorithm for solving sub-problems is $O(sk^2_{max} * 2^{k_{max}-2})$ where $s$ is the number of sub-problems and $k_{max}$ is size of the largest of them. So, the total time complexity is = $O(n^{\sigma+3} + n* m^{\sigma+1}_{max} + sk^2_{max} * 2^{k_{max}-2})*CT$ where $CT$ is the time complexity for a single CI-test. However, the time complexity for solving the complete problem with PC is $O(n^22^{n-2})*CT$. Since the parameter $k_{max}$ used in {\sf Dsep-CP} is much lower than $n$, {\sf Dsep-CP} experiences notably reduced time complexity than PC.
	\vspace{-2mm}
	\section{Empirical results}
	In this section, we empirically evaluate {\sf Dsep-CP} compared to three different state-of-the-art causal learning algorithms in terms of solution quality and scalability. We run our experiments on different simulated causal structures and on eight real-world causal networks. We show that {\sf Dsep-CP} performs better than CAPA ~\cite{zhang2019recursively}, SADA ~\cite{cai2013sada} and CP ~\cite{yan2020effective} in terms of scalability while yielding solutions of similar or better quality. To conduct all these experiments, we run parallel instances of all the competing algorithms in an Intel Xeon 20 Core machine with 92GB RAM (code: \href{https://github.com/softsys4ai/Dsep-CP}{github.com/softsys4ai/{\sf Dsep-CP}.})
	
	In all the experiments, we use the linear non-Gaussian model for sample data generation from any causal structures, according to~\cite{cai2013sada}. Briefly, for each node $i$ in topologically sorted order, we generate a linear function $v_i = \sum_{j\epsilon P_i} w_{ji} v_j + r\varepsilon_i$ and evaluate it for a number of times called sample size to produce the sample set. Here $P_i$ is the set of parents of node $i$, $w_{ji}$ is a weight denoting the effect of $v_j$ on $v_i$, $\varepsilon_i$ is the non-Gaussian noise term and $r = 0.3$ is a constant value denoting the effect of the noise term on $v_i$. During the generation of these linear functions, for each node $i$, we ensure that the arithmetic mean of $\varepsilon_i$ is 0 and the variance is 1, as well as $\sum_{j\epsilon P_i} w_{ji} = 1$. Finally, we shift all the values of $v_i$ by their arithmetic mean and scale them by their variance (i.e., normalize the values) such that the new arithmetic mean and the new variance of these values also become 0 and 1, respectively. We set the maximum size of conditioning sets for CI-tests to 3. We do this because most practical causal networks are sparse and this threshold is sufficient to discover them. Also, size > 3 may produce type II errors ~\cite{zhang2019recursively}. Moreover, we terminate the causal partitioning process when the size of the corresponding subset is smaller than or equal to $\max(\lfloor N/10\rfloor ,3)$, where $N$ is the total number of nodes. Finally, each algorithm deploys the PC algorithm to discover causal graphs from the smallest sub-problems.
	
	In our experiments, we employ the algorithms on the generated sample data to find causal skeletons (without directions) instead of finding the causal structures (with directions). We do this based on the recommendation in ~\cite{yan2020effective}.
	However, we can use V-structure based methods ~\cite{cai2011bassum, cai2013causal}, Additive Noise Models ~\cite{peters2011causal}, Information Geometric Causal Inference ~\cite{budhathoki2018origo}, etc. to acquire the directions of causal skeletons. To report each of the results, we run each algorithm 20 times for a different number of nodes (or samples). In so doing, we consider the arithmetic mean value over all corresponding runs and 95\% confidence interval (for error bars). The confidence interval is based on the standard error of mean and z-value = 1.96. All our reported results are significant for p-value < 0.05.
	
	\begin{figure}[t]
		\begin{subfigure}[h]{.495\linewidth}
			\centering
			\includegraphics[width=\linewidth]{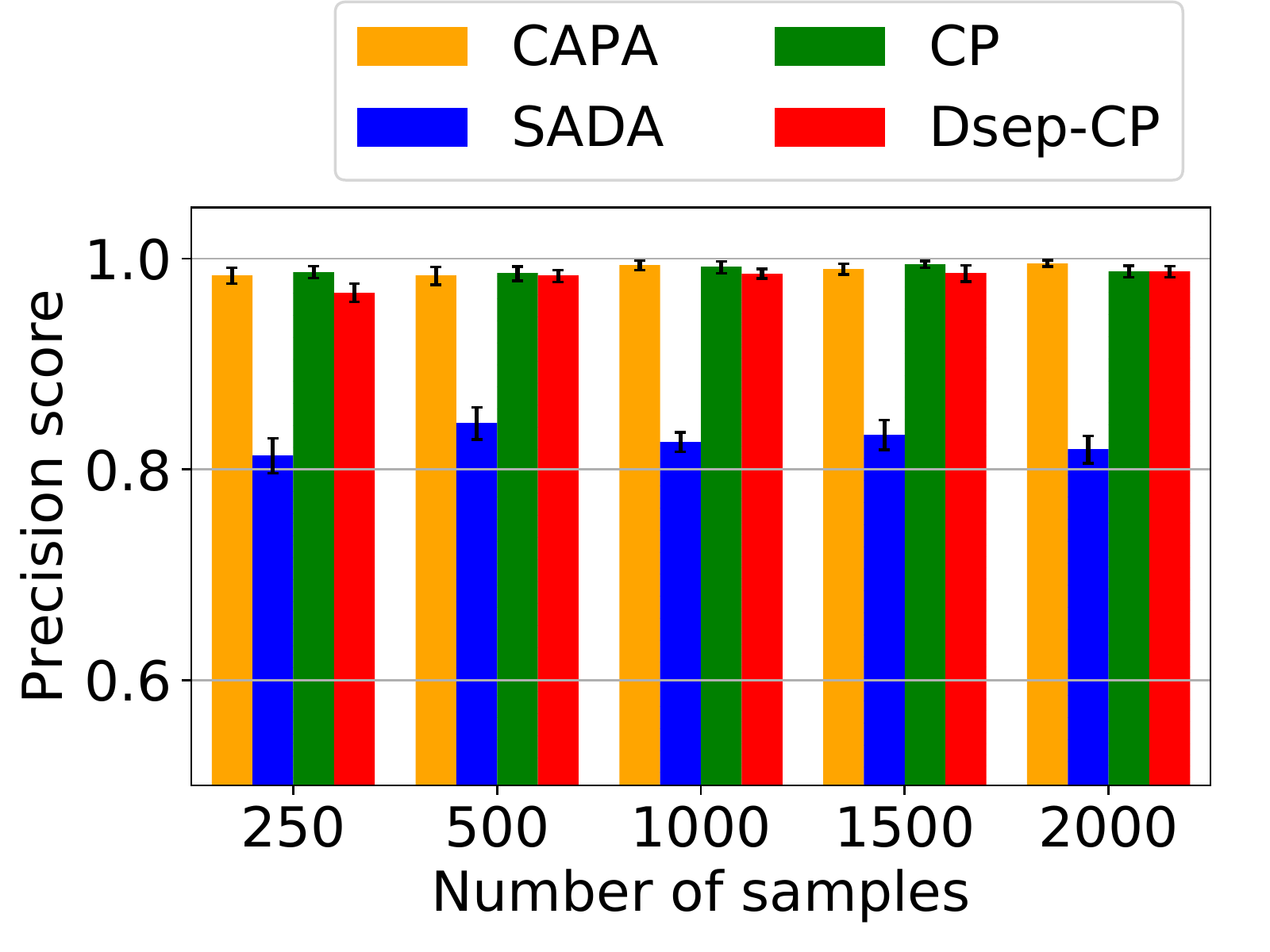}
			\vspace{-2mm}
			\caption{Precision score}
			\label{fig:simulated_precision}
			\Description{}
		\end{subfigure}
		\begin{subfigure}[h]{.495\linewidth}
			\centering
			\includegraphics[width=\linewidth]{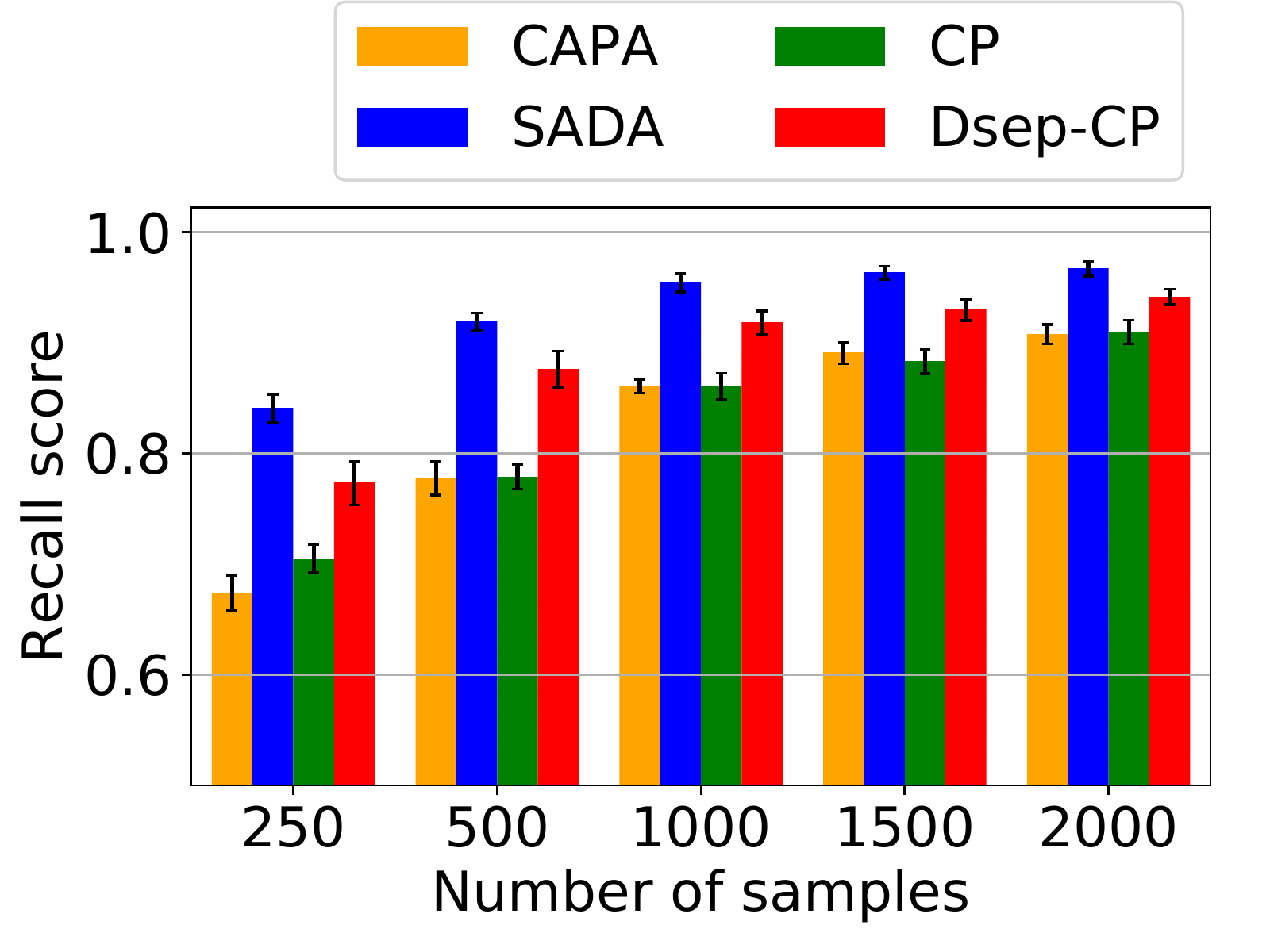}
			\vspace{-2mm}
			\caption{Recall score}
			\label{fig:simulated_recall}
			\Description{}
		\end{subfigure}
		\begin{subfigure}[h]{.495\linewidth}
			\centering
			\includegraphics[width=\linewidth]{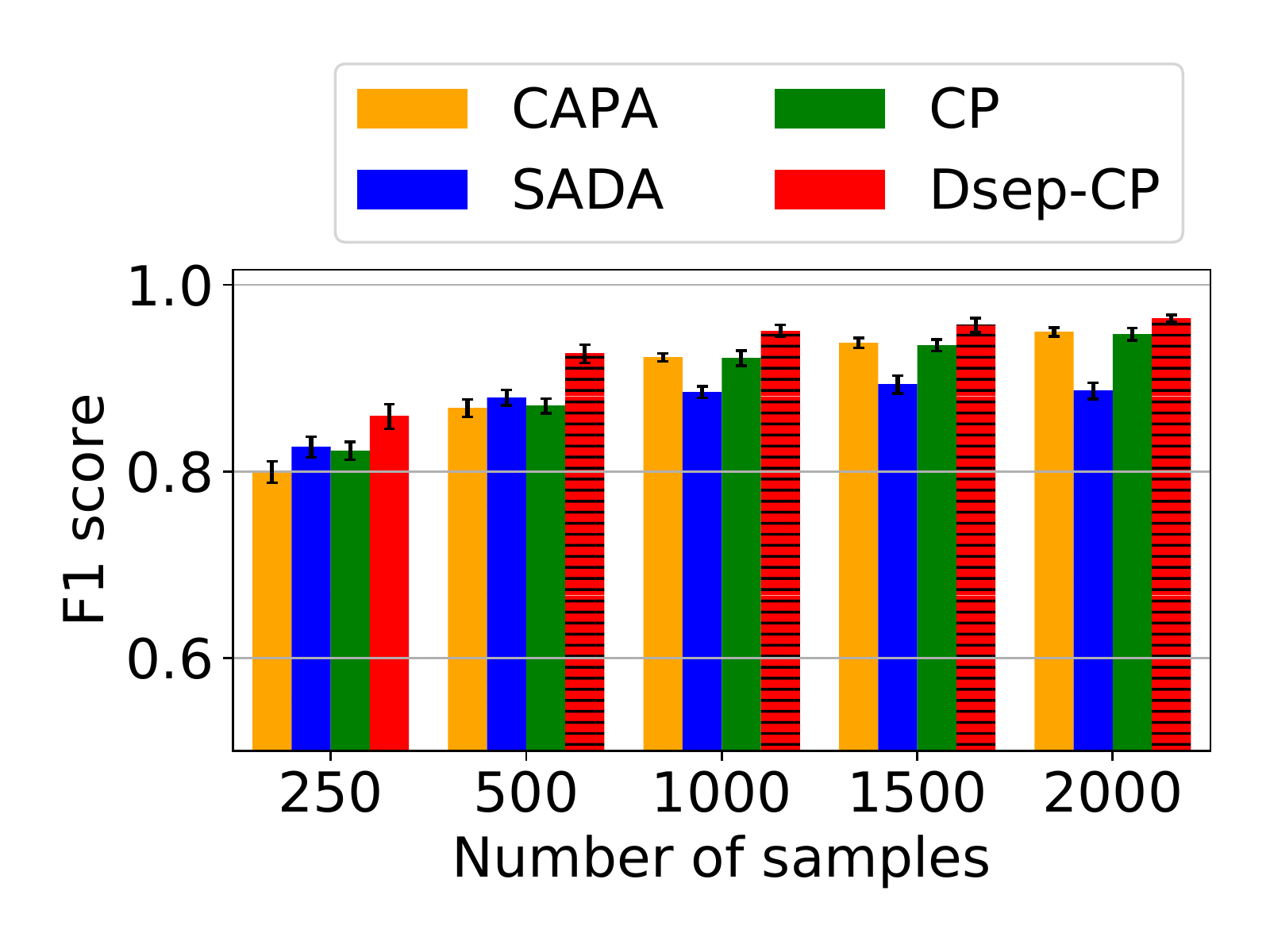}
			\caption{F1 score}
			\label{fig:simulated_f1}
			\Description{}
		\end{subfigure}
		\begin{subfigure}[h]{.495\linewidth}
			\centering
			\includegraphics[width=\linewidth]{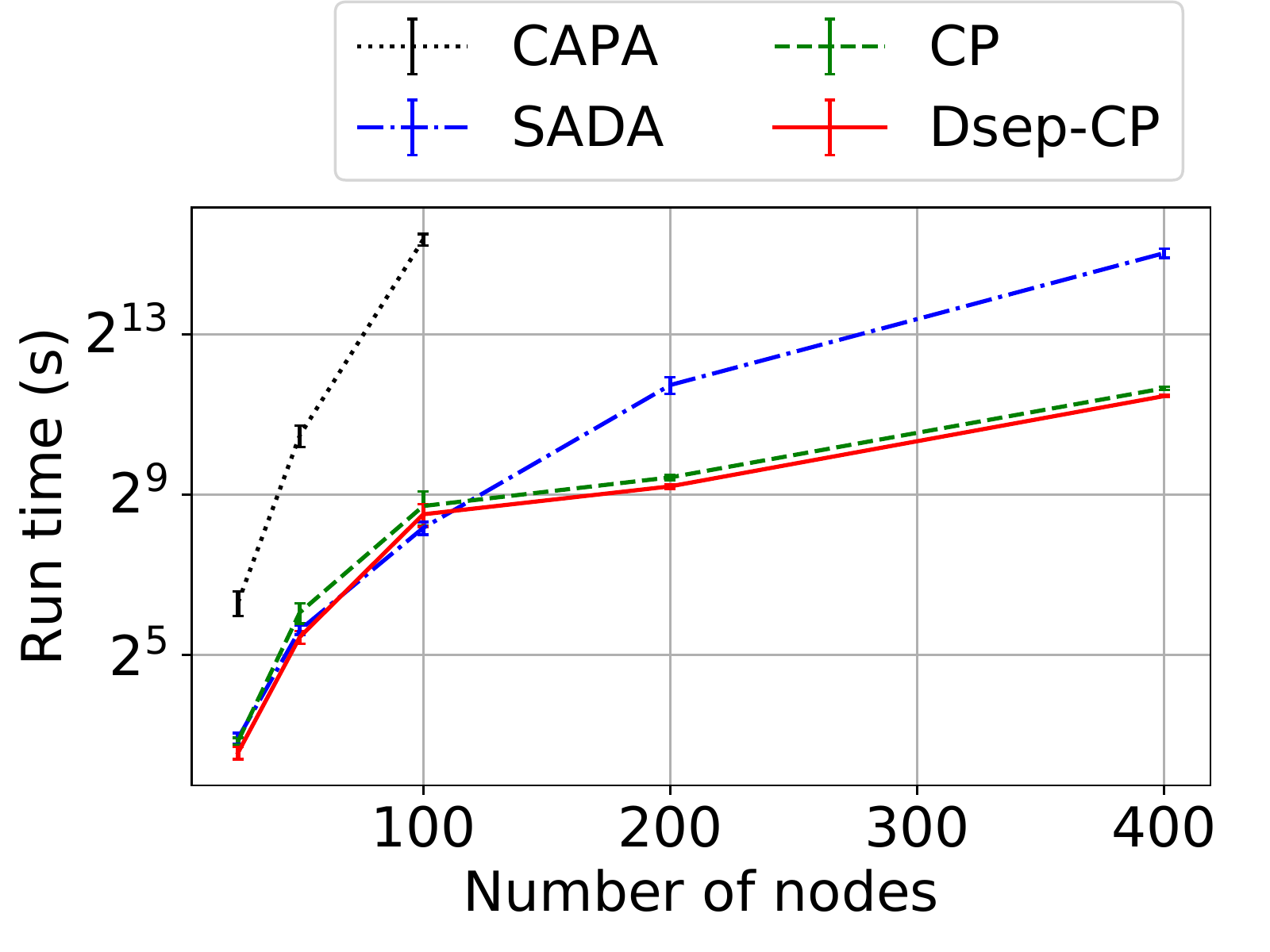}
			\caption{Run-time}
			\label{fig:simulated_runtime}
			\Description{}
		\end{subfigure}
		\begin{subfigure}[h]{.495\linewidth}
			\centering
			\includegraphics[width=\linewidth]{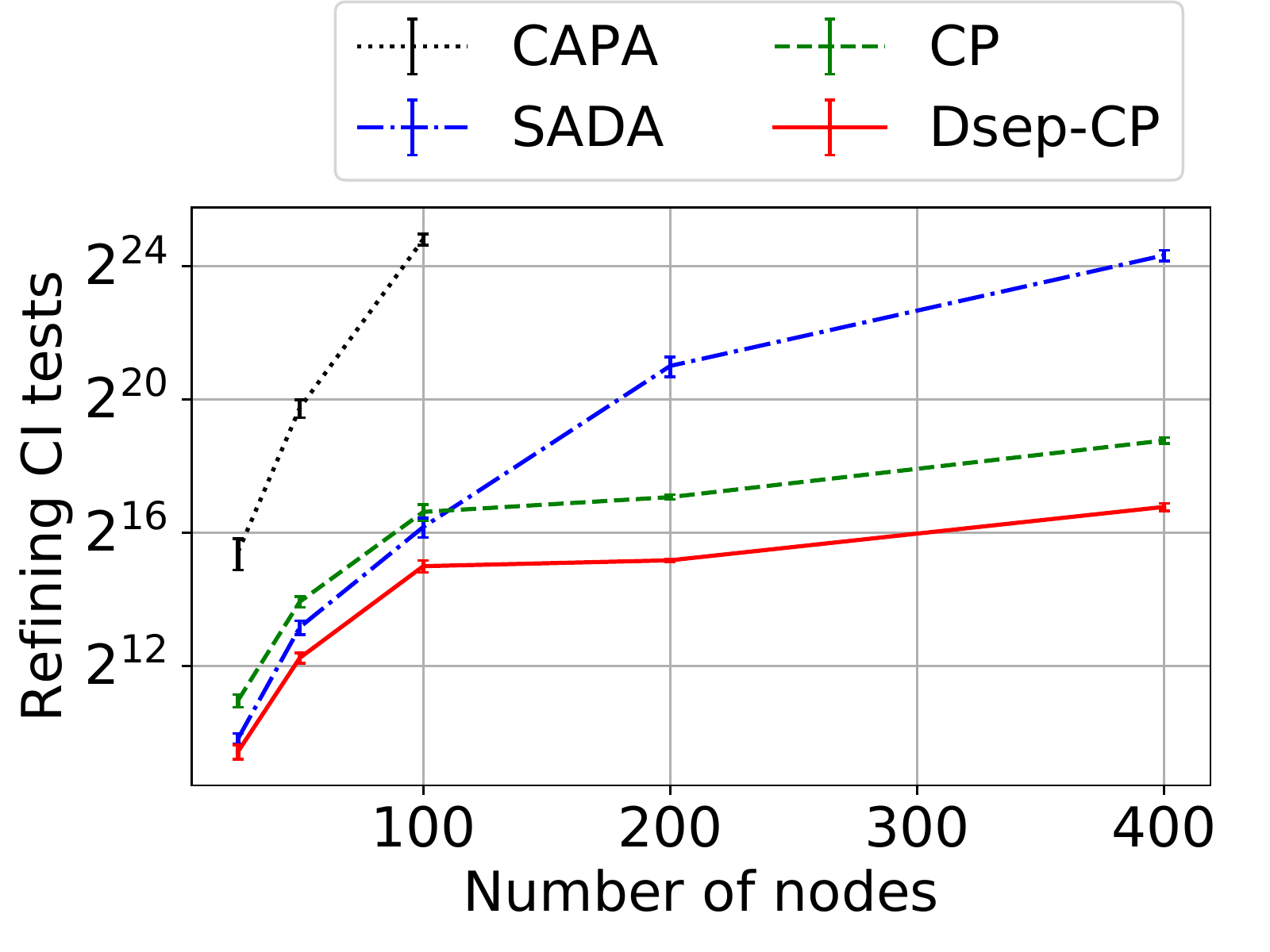}
			\caption{Number of refining CI-tests}
			\label{fig:simulated_ccitest}
			\Description{}
		\end{subfigure}
		\begin{subfigure}[h]{.495\linewidth}
			\centering
			\includegraphics[width=\linewidth]{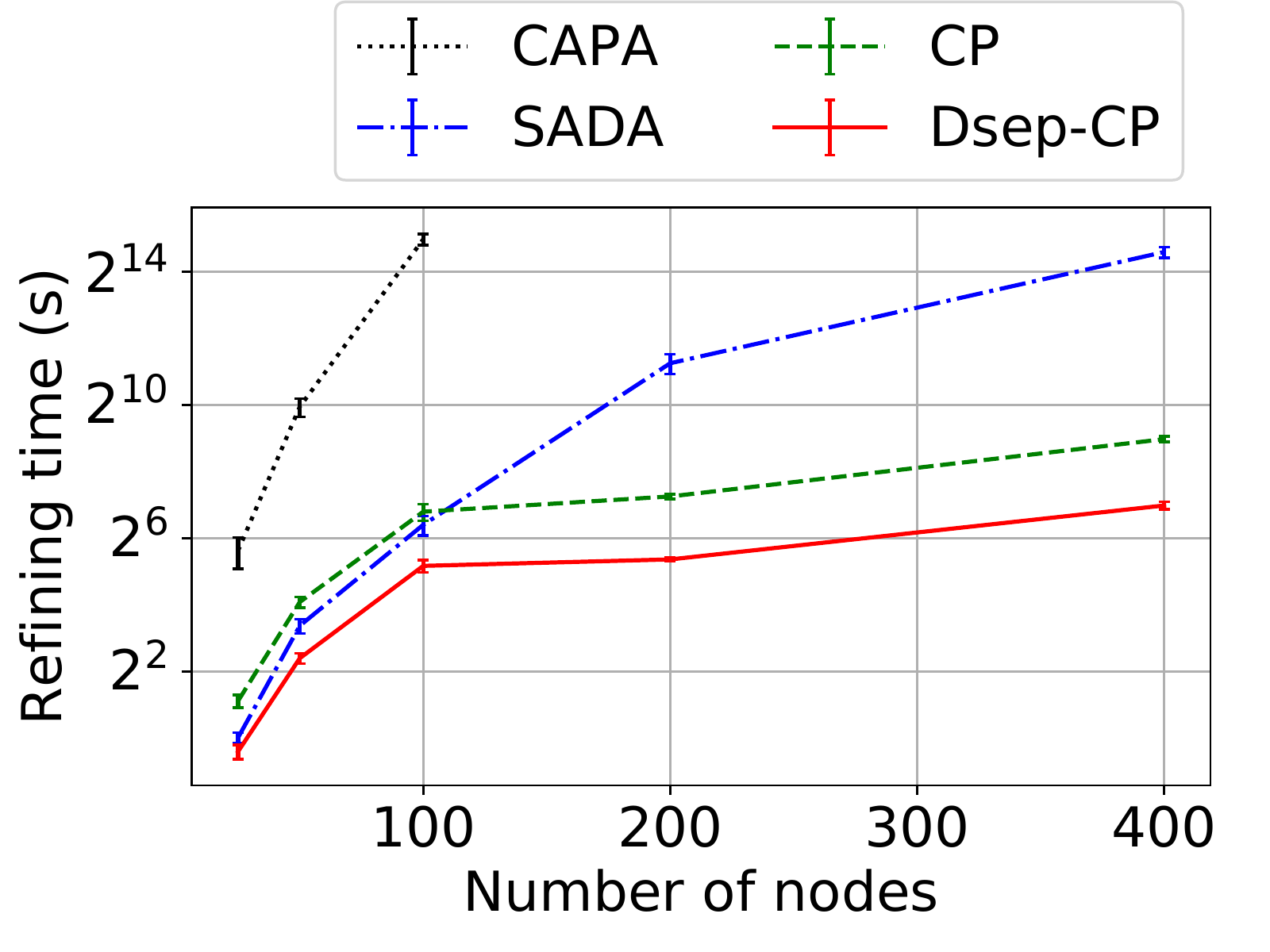}
			\caption{Refining time}
			\label{fig:simulated_reftime}
			\Description{}
		\end{subfigure}
		\caption{Results of experiments on simulated structures}
	\end{figure}
	
	\begin{figure*}[h]
		\begin{subfigure}[h]{.3\textwidth}
			\centering
			\includegraphics[width=\linewidth]{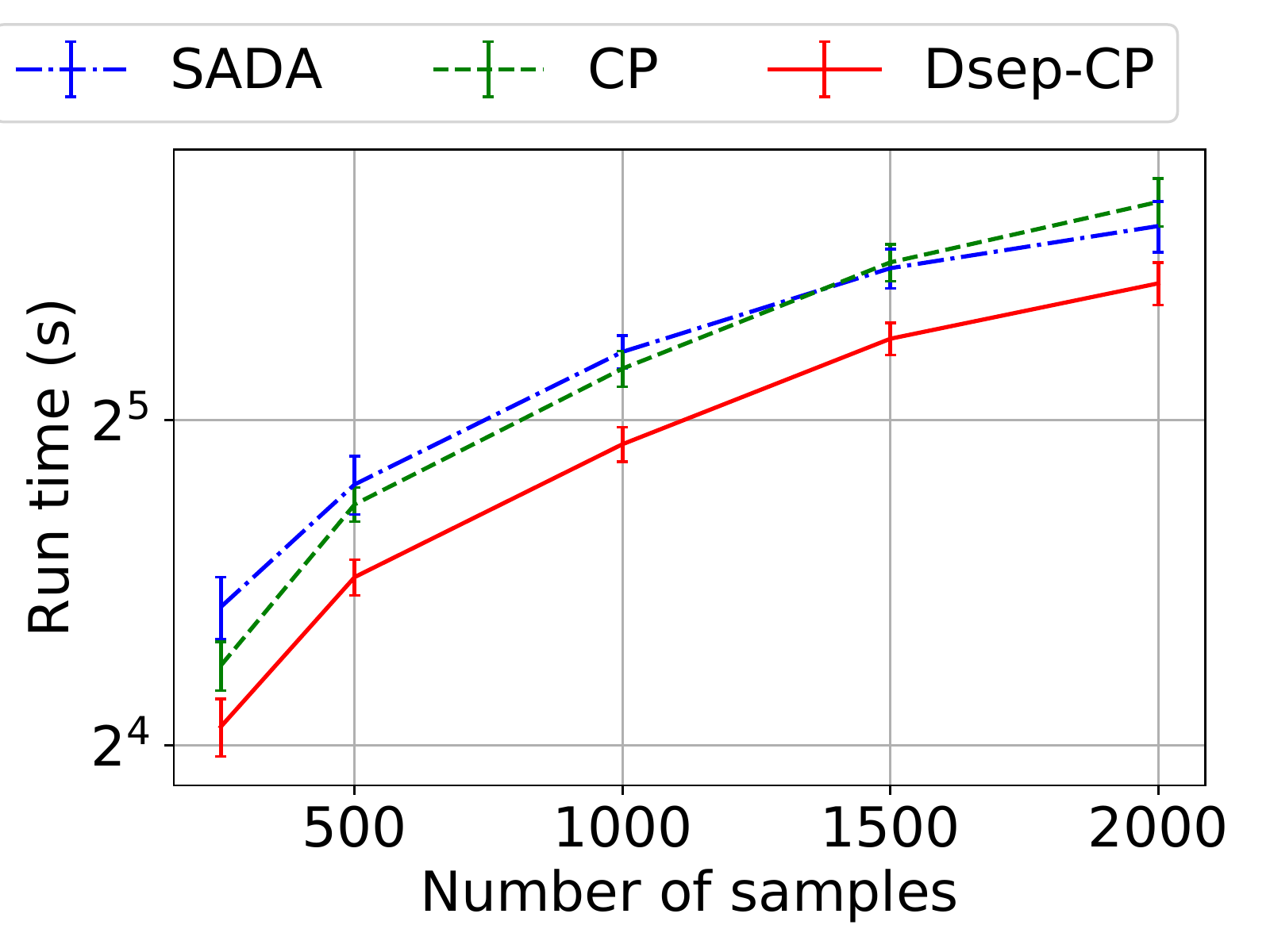}
			\caption{Run-time}
			\label{fig:mildew_runtime}
			\Description{}
		\end{subfigure}
		\begin{subfigure}[h]{.3\textwidth}
			\centering
			\includegraphics[width=\linewidth]{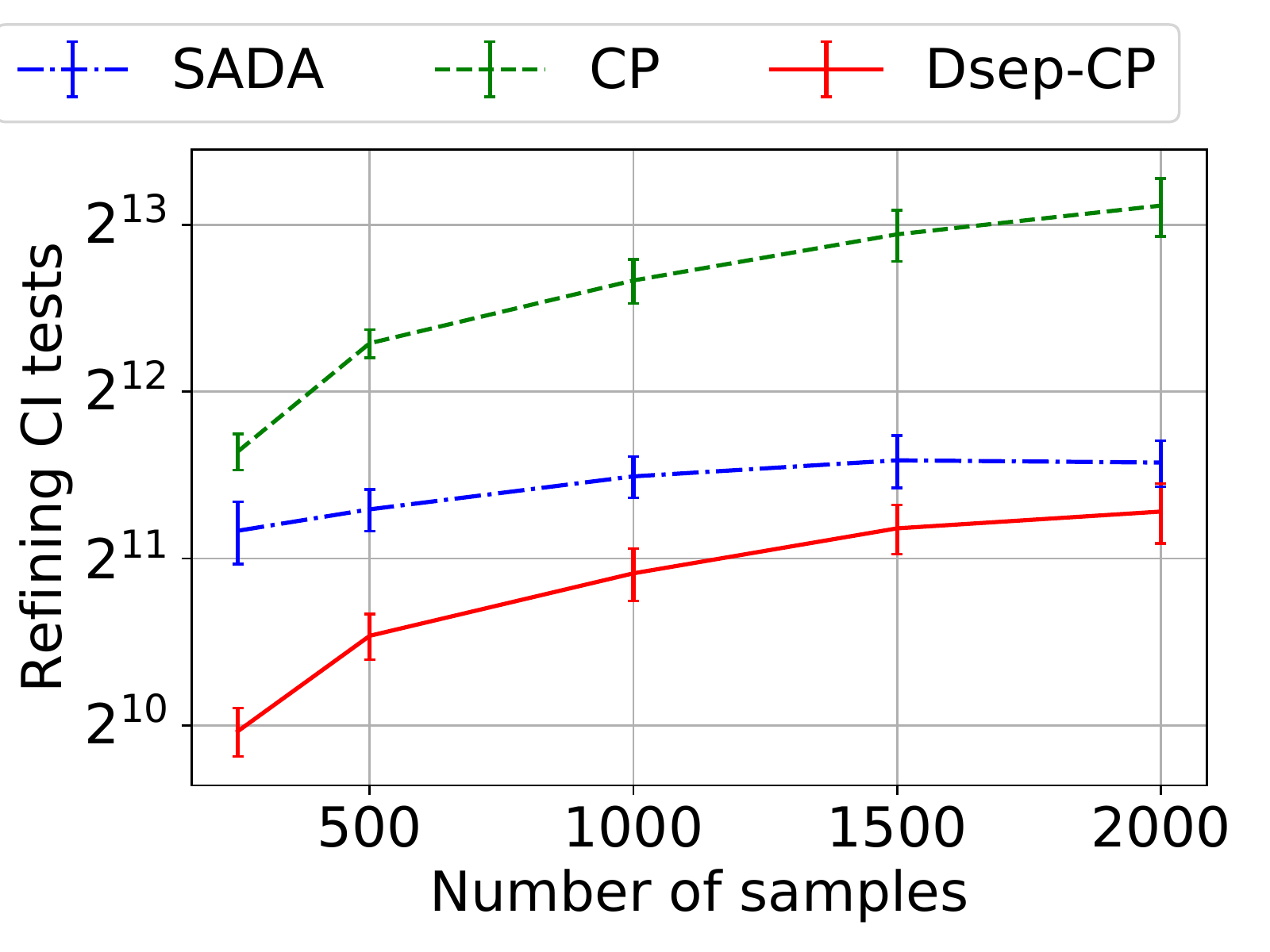}
			\caption{Number of refining CI-tests}
			\label{fig:mildew_ccitest}
			\Description{}
		\end{subfigure}
		\begin{subfigure}[h]{.3\textwidth}
			\centering
			\includegraphics[width=\linewidth]{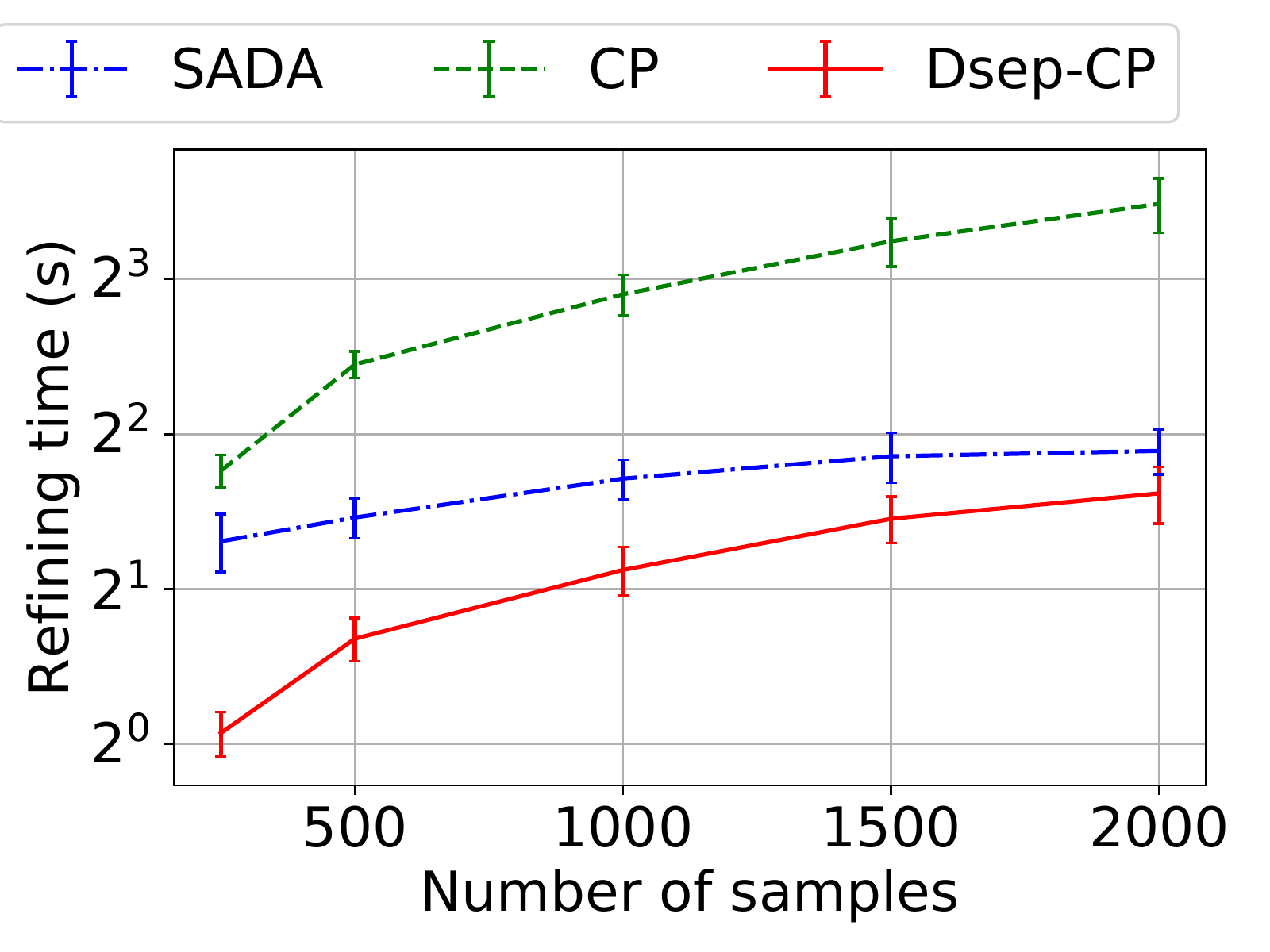}
			\caption{Refining time}
			\label{fig:mildew_reftime}
			\Description{}
		\end{subfigure}
		\vspace{-4mm}
		\caption{Results of experiments on a real-world causal structure (Mildew)}
	\end{figure*}
	\vspace{-3mm}
	\begin{table*}[t]
		\caption{Solution quality of algorithms on the real-world causal structures with sample size = 500 (best values in bold font).\vspace{-2mm}}
		\label{tab:realNetworksQuality}
		\begin{tabular}{|l|r|r||r|r|r||r|r|r||r|r|r|} \toprule
			Network & Average & Maximum &
			\multicolumn{3}{c||}{F1 score} &
			\multicolumn{3}{c||}{Recall score} &
			\multicolumn{3}{c|}{Precision score} \\
			name & degree & in-degree & SADA & CP & {\sf Dsep-CP} & SADA & CP & {\sf Dsep-CP} & SADA & CP & {\sf Dsep-CP} \\ \midrule
            Asia       & 2    & 2 & 0.620294 & \textbf{0.910476} & 0.908452 & 0.600000 & 0.837500 & \textbf{0.843750} & 0.644048 & \textbf{1.000000} & 0.987500 \\ \hline
            Sachs      & 3.09 & 3 & 0.985714 & 0.967023 & \textbf{0.968538} & \textbf{0.997059} & 0.947059 & 0.950000 & 0.975000 & \textbf{0.988235} & \textbf{0.988235} \\ \hline
            Water      & 4.12 & 5 & 0.425022 & 0.766123 & \textbf{0.787150} & 0.337879 & 0.635606 & \textbf{0.679545} & 0.576877 & \textbf{0.964537} & 0.936064 \\ \hline
            Mildew     & 2.63 & 3 & 0.641518 & 0.839170 & \textbf{0.878156} & 0.660870 & 0.740217 & \textbf{0.817391} & 0.624779 & \textbf{0.969078} & 0.949457 \\ \hline
            Hepar2     & 3.51 & 6 & 0.507614 & 0.580315 & \textbf{0.637615} & 0.426016 & 0.479675 & \textbf{0.560976} & 0.629615 & 0.735451 & \textbf{0.739092} \\ \hline
            Win95pts   & 2.95 & 7 & 0.533644 & 0.862308 & \textbf{0.898965} & 0.463839 & 0.785714 & \textbf{0.856696} & 0.631409 & \textbf{0.956089} & 0.946412 \\ \hline
            Andes      & 3.03 & 6 & 0.354775 & 0.917100 & \textbf{0.929056} & 0.304241 & 0.900066 & \textbf{0.943294} & 0.427975 & \textbf{0.934940} & 0.915381 \\ \hline
            Pigs       & 2.68 & 2 & 0.692010 & 0.893390 & \textbf{0.911480} & 0.785811 & 0.855405 & \textbf{0.928716} & 0.619149 & \textbf{0.935103} & 0.895006 \\ 
			\bottomrule
		\end{tabular}
	\end{table*}
	\vspace{-5mm}
	\begin{table*}[t]
		\caption{Scalability of algorithms on the real-world causal structures with sample size = 500 (best values in bold font).}
		\label{tab:realNetworksScalability}
		\begin{tabular}{|l|r|r||r|r|r||r|r|r||r|r|r|} \toprule
			Network & Number & Number &
			\multicolumn{3}{c||}{Run-time (s)} &
			\multicolumn{3}{c||}{Number of refining CI-tests} &
			\multicolumn{3}{c|}{Refining time (s)} \\
			name & of nodes & of arcs & SADA & CP & {\sf Dsep-CP} & SADA & CP & {\sf Dsep-CP} & SADA & CP & {\sf Dsep-CP} \\ \midrule
            Asia       & 8  & 8   & 0.45   & 0.26   & \textbf{0.23}   & 38.45   & 36.80    & \textbf{10.60}   & 0.04 & 0.04  & \textbf{0.01} \\ \hline
            Sachs      & 11 & 17  & 1.41   & 1.20   & \textbf{0.84}   & 98.00   & 343.15   & \textbf{8.00}    & 0.10 & 0.37  & \textbf{0.01} \\ \hline
            Water      & 32 & 66  & 30.99  & 27.43  & \textbf{20.21}  & 1787.85 & 8262.15  & \textbf{1660.80} & 1.93 & 9.02  & \textbf{1.79} \\ \hline
            Mildew     & 35 & 46  & 27.90  & 26.73  & \textbf{22.90}  & 2512.90 & 5013.00  & \textbf{1485.30} & 2.75 & 5.46  & \textbf{1.60} \\ \hline
            Hepar2     & 70 & 123 & 142.12 & 146.85 & \textbf{125.23} & \textbf{4685.05} & 25247.75 & 5526.60 & \textbf{5.23} & 27.88 & 6.06 \\ \hline
            Win95pts   & 76 & 112 & 99.20  & 129.49 & \textbf{97.90}  & 8968.60 & 36720.70 & \textbf{7597.15} & 9.83 & 40.05 & \textbf{8.21} \\ \hline
            Andes      & 223 & 338  & 3446.13  & 2465.09  & \textbf{2185.73}  & 87477.50   & 294446.89  & \textbf{42588.89}  & 102.28  & 337.62  & \textbf{48.43}  \\ \hline
            Pigs       & 441 & 592  & 17952.85 & 8729.67  & \textbf{7560.66}  & 1120269.20 & 1166059.60 & \textbf{223835.60} & 1338.21 & 1354.75 & \textbf{258.20} \\ 
			\bottomrule
		\end{tabular}
	\end{table*}
	
\vspace{3.9mm}
\subsection{Experiments on Simulated Structures}

In this group of experiments, we compare CAPA, SADA, CP and {\sf Dsep-CP} on linear non-Gaussian sample data generated by randomly generated causal structures. During the generation of each node in true causal structure, we select an average of 1.5 previously generated nodes as the child of the current node, as recommended in ~\cite{zhang2019recursively}. We first run all the algorithms using different sample sizes \{250, 500, 1000, 1500, 2000\}, following ~\cite{yan2020effective}, with fixed dimension (i.e., we choose 50 nodes to save time for all algorithms). The results are presented in Figures ~\ref{fig:simulated_precision} - ~\ref{fig:simulated_f1}. For all algorithms, these figures show comparative precision (i.e., indicator of the amount of redundant edges), recall (i.e., indicator of the amount of missing edges) and F1 scores, respectively (see \cite{yan2020effective} for more detail). It can be observed from these figures that {\sf Dsep-CP} produces solutions of similar quality compared to the benchmarking algorithms for varying sample sizes on simulated networks. This result is significant because it illustrates that our algorithm does not sacrifice the solution quality compared to its competitors.

	We also run CAPA, SADA, CP and {\sf Dsep-CP} using the same simulated model over varying dimensional networks \{25, 50, 100, 200, 400\} as done in ~\cite{zhang2019recursively}, with fixed (500) sample size. However, the run-time of CAPA is prohibitively so expensive that we are not able to run this algorithm for simulated networks of a size larger than 100 \footnote{Note that, we are particularly concerned about the larger networks, as the smaller ones are trivial to handle by any of the algorithms.}. The results are presented in Figures ~\ref{fig:simulated_runtime} - ~\ref{fig:simulated_reftime}. These figures show comparative run-time (in seconds), numbers of refining CI-tests and refining time (in seconds), respectively.
    We define the number of CI-tests and execution time required for the refinement purpose of all the algorithms as the number of refining CI-tests and refining time, respectively. These parameters help us to investigate the contribution of our improved refinement phase.
    
	It can be observed from figures ~\ref{fig:simulated_runtime} - ~\ref{fig:simulated_reftime} that for higher dimensional networks, {\sf Dsep-CP} scales better than the competing algorithms. To be precise, for 200-400 nodes, {\sf Dsep-CP} runs 83-92\% faster than SADA and 13-14\% faster than CP while reducing both refining time and refining CI-tests by 98-99\% compared to SADA and by 73-75\% compared to CP.
	This improvement is possible due to its ability to detect and remove false edges with less number of CI-tests during its refinement procedure.
	We also see from these figures that for higher dimensional networks, CAPA is not scalable. Notably, from Figures ~\ref{fig:simulated_precision} - ~\ref{fig:simulated_f1}, we see that the quality of solutions produced by CAPA is similar to some of the other algorithms. Considering this, coupled with its scalability issue, we opt not to include CAPA in the remaining experiments.
	\vspace{-3mm}
	\subsection{Experiments on Real-World Structures}
	In this group of experiments, we compare SADA, CP and {\sf Dsep-CP} on linear non-Gaussian sample data generated by some real-world causal structures. We select these causal structures because they cover a variety of applications, including causal inference (Asia), protein signaling network (Sachs), waste water treatment (Water), disease risk forecasting (Mildew), diagnosis of liver disorders (Hepar2), printer troubleshooting (Win95pts), intelligent tutoring system (Andes) and the pedigree of breeding pigs (Pigs). Moreover, we particularly select these networks to cover different sizes of networks. Table ~\ref{tab:realNetworksQuality} and ~\ref{tab:realNetworksScalability} show the names and some structural information about each network, in their first three columns.
	
	We run SADA, CP and {\sf Dsep-CP} using different sample sizes \{250, 500, 1000, 1500, 2000\} for each of the real-world causal structures following ~\cite{yan2020effective}. However, for two very large networks i.e. Andes and Pigs, we cannot run all algorithms 20 times. Instead, we run all algorithms on Andes and Pigs networks 18 and 5 times, respectively. The results for Mildew network is presented in Figures ~\ref{fig:mildew_runtime} - ~\ref{fig:mildew_reftime}. These figures show comparative run-times (in seconds), numbers of refining CI-tests and refining times (in seconds), respectively. From these figures, we see that {\sf Dsep-CP} runs faster than the other algorithms while producing solutions of similar quality for varying sample sizes for real-world networks. Notably, we observe similarities between the figures of number of refining CI-tests and refining time (in Figure~\ref{fig:simulated_ccitest} and~\ref{fig:simulated_reftime}, or Figure~\ref{fig:mildew_ccitest} and~\ref{fig:mildew_reftime}) since they are proportionate to each other. Additionally, in Figure~\ref{fig:mildew_ccitest} - ~\ref{fig:mildew_reftime}, SADA gets closer to {\sf Dsep-CP} with increasing number of samples, but it is outperformed by {\sf Dsep-CP} due to its solution quality.

	We observe comparable results by running our experiments on other real-world networks. Due to lack of space, we present these results for a fixed sample size (i.e. 500) in Table~\ref{tab:realNetworksQuality} and Table ~\ref{tab:realNetworksScalability}. Table ~\ref{tab:realNetworksQuality} shows the solution quality of the three algorithms in terms of F1 scores, recall scores and precision scores. From this table, we again see that {\sf Dsep-CP} yields solutions of similar quality compared to SADA and CP. Table ~\ref{tab:realNetworksScalability} shows the performance of the three algorithms in terms of total run-time, numbers of refining CI-tests and refining times. We also observe from the table that for very large real-world networks (200+ nodes), {\sf Dsep-CP} runs around 37-58\% faster than SADA and around 11-13\% faster than CP while reducing both refining time and CI-tests by 51-81\% compared to SADA and by 81-86\% compared to CP. So, we can conclude that {\sf Dsep-CP} is more applicable to causal discovery in high-dimensional cases than the state-of-the-art algorithms.

	\balance
	\vspace{-1mm}
	\section{Conclusions and Future Work}
This paper introduces a recursive causal structure learning algorithm, {\sf Dsep-CP}, to support effective and efficient causal discovery for large sets of variables. To incorporate our recursive method, we present an improved refinement mechanism that can reduce the algorithm's execution time without compromising the solution quality. In our theoretical section, we prove the correctness of {\sf Dsep-CP}. Finally, our extensive empirical observation illustrates that {\sf Dsep-CP} outperforms the state-of-the-art algorithms in both synthetic and real-world structures. To be precise, {\sf Dsep-CP} runs up to 92\% faster than SADA and up to 14\% faster than the CP algorithm. In the future, we intend to investigate whether this algorithm can be improved for faster causality discovery while increasing the recall value (i.e., recovering the falsely removed edges) by analyzing the characteristics of CI-tests.
	
	\begin{acks}

This research is mainly supported by the ICT Innovation Fund of Bangladesh Government. We also acknowledge the use of the BdREN High Performance Computing Facility. Moreover, the work done by Pooyan Jamshidi is supported by NASA (RASPERRY-SI Grant No. 80NSSC20K1720) and NSF (SmartSight Award 2007202).
\end{acks}
	
	
\bibliographystyle{ACM-Reference-Format}
\bibliography{main}
\end{document}